\documentclass[letterpaper, 10 pt, conference]{ieeeconf}  

\IEEEoverridecommandlockouts                              

\usepackage{xspace}
\newcommand{\rpb}{robust, perception based controller\xspace} 

\usepackage{rebeccaNotes}
\newcommand{\comment}[1]{}  

\title{\LARGE \bf
Robust, Perception Based Control with Quadrotors
}
\author{Laura Jarin-Lipschitz$^{\dagger}$, Rebecca Li$^{\dagger}$, Ty Nguyen, Vijay Kumar, and Nikolai Matni
\thanks{$^\dagger$ These authors contributed equally to the work}%
\thanks{The authors are a part of the GRASP Lab, University of Pennsylvania, PA, 19104, USA {\tt\footnotesize \{laurajar, robot, kumar, tynguyen, nmatni\}@seas.upenn.edu.}}%
}

\begin{document}
\maketitle
\begin{abstract}
Traditionally, controllers and state estimators in robotic systems are designed independently. Controllers are often designed assuming perfect state estimation. However, state estimation methods such as Visual Inertial Odometry (VIO) drift over time and can cause the system to misbehave. While state estimation error can be corrected with the aid of GPS or motion capture, these complementary sensors are not always available or reliable. Recent work has shown that this issue can be dealt with by synthesizing robust controllers using a data-driven characterization of the perception error, and can bound the system's response to state estimation error using a robustness constraint. We investigate the application of this robust perception-based approach to a quadrotor model using VIO for state estimation and demonstrate the benefits and drawbacks of using this technique in simulation and hardware. Additionally, to make tuning easier, we introduce a new cost function to use in the control synthesis which allows one to take an existing controller and "robustify" it. To the best of our knowledge, this is the first \textit{robust} perception-based controller implemented in real hardware, as well as one utilizing a data-driven perception model. We believe this as an important step towards safe, robust robots that explicitly account for the inherent dependence between perception and control.
\end{abstract}

\section{INTRODUCTION}
In real-world robotics with size, weight, and power constraints, a vision sensor is a cheap and lightweight sensor solution to localize a robot in its environment.
In particular, vision-based localization using Visual-Inertial Odometry (VIO) has been an important technique in bringing multi-rotor robots out of the lab and into real-world environments \cite{Loianno2016}. However, visual perception systems can produce state estimates that drift over time, accumulating error that ultimately limits the capability of the robotic system.  Poor lighting conditions and a lack of features to track in the environment can make visual state estimation even worse.

While many robotic systems seek to solve this using near-ideal sensors such as motion capture or GPS, this is often impractical or impossible. For lightweight or low-cost robotic systems, adding GPS or other ground truth localization sensors might not be possible. Additionally, there are many cases such as GPS-denied environments or underground environments where robots may not get GPS signal updates in order to correct state estimation error growth. Localization using motion capture systems such as VICON is only possible in lab and indoor environments, significantly limiting use in the real world. 

Historically, roboticists have assumed the state estimation of their system to be "good enough," and proceeded to perform control and planning while assuming perfect state information.
Theoretically, this is supported by the separation principle \cite{separation_principle}, which says that for certain linear stochastic systems, optimal control and optimal state estimation can be decoupled and still remain optimal. The assumptions needed for these guarantees (e.g. quadratic cost, unbiased Gaussian noise) are rarely, if ever satisfied by real-world vision-based systems. Instead, control actions taken by the system can significantly impact the quality and content of the images used for state estimation. In this work, we explicitly account for the interplay between perception and control, and furthermore leverage the controller to reduce perception error in a robotic system.


The approach we take in this work is to make the system robust to state estimation error through the use of a \rpb. We follow the approach outlined in \cite{Dean2019a} to incorporate a data-driven perception error model into an optimal controller synthesis procedure.
By utilizing this procedure, we can guarantee that the perception error will remain bounded given a well-characterized perception map.
Specifically, we implement the \rpb on a quadrotor, a platform where VIO is preferable to other state estimators due to its size, weight, and cost constraints.
We demonstrate that for a quadrotor, the state estimation and tracking for the \rpb is considerably better than an $L_1$ controller, and matches that of a tuned PD controller on typical conditions. We also demonstrate that the \rpb remains robust to perturbations in the environment such as weaker ambient light and fewer texture features while the PD control suffers.
In short, our contributions include:
\begin{itemize}
    \item Introduction and implementation of a \rpb with a data-driven error model on a quadrotor with VIO, in simulation and reality. To the best of our knowledge, we are the first to implement a \textit{robust}, perception-based controller on a hardware system.
    \item Introduction of a new cost function, the \textit{imitation cost}, which allows the roboticist to use an existing controller and "robustify" it.
    \item Evaluation of the \rpb with a data-driven error model using traditional quadratic cost functions and the imitation cost function, in comparison to Proportional-Derivative (PD) and quadratic cost L1 optimal controllers. We evaluate these controllers in ideal perception and degraded perception conditions.
\end{itemize}

\section{RELATED WORK}
Controllers that incorporate perception into control and planning have been an important part of enabling robotic systems to move into unstructured environments. For quadrotor platforms, Visual-Inertial Odometry has been a popular technique to fuse vision with IMU measurements for accurate state estimation \cite{Loianno2016},  \cite{Sun2018}. These utilize Kalman filter based estimators to produce good state estimates. However, state estimation error with these algorithms grows \comment{grow to grows}significantly over time. Simultaneous Localization and Mapping (SLAM) is a tool used to match current perception measurements to a constructed map, and is used widely in UAV platforms \cite{big_slam}, \cite{orbslam}. Unfortunately, SLAM has extremely high computational and memory load, which makes it unsuitable for small or low-cost platforms.
\comment{Maybe better to include vision-based controller papers such as visual servoing. For example (I've just googled:
https://ieeexplore.ieee.org/document/8746787
https://www.sciencedirect.com/science/article/pii/S1474667016398597
https://journals.sagepub.com/doi/abs/10.1177/0142331217751599
Giusseppe also had some papers too.
}

Another approach is to explicitly consider perception in the control or planning. In \cite{PAMPC}, a Model Predictive Controller (MPC) controller is presented which, in addition to having trajectory following, \comment{added ,}also has perception objectives. Perception-aware planning is yet another approach, which can avoid parts of the space where localization may be lost \cite{ichter2017perceptionaware}. Visual servoing, or explicitly controlling given vision input, has incorporated robustness in \cite{robust_vis_servo} and \cite{vis_serv_sim} by explicitly take into account model error, but not perception error in their models.

More accurate perception models can be achieved through learning. In \cite{bansal2019combining}, a rover robot was able to use a learned perception model to navigate safely through a novel environment. Racing drones with learning perception systems were also demonstrated by \cite{kaufmann2018beauty}. However, none of these systems provide any control-theoretic guarantees, nor do they attempt to bound the perception or tracking error of the system. In contrast, we define the explicit consideration and presence of guarantees to constitute robust control, which is extremely important in safety-critical applications. In \cite{Dean2019a}, the authors introduce a \rpb, but only demonstrate it on a relatively simple double integrator system. In this work, we seek to get much closer to bridging the gap between the simplistic models that are often used in robotics, and real, complex systems by incorporating robust guarantees informed by data-driven perception models.

\section{Problem Formulation}
\subsection{Notation}
We use lower case letters $x$ to denote vectors, and capital letters $A$ to denote matrices. $x_k$ denotes the discrete signal $x$ at time $k$, while $x_{0:k}$ denotes the discrete signal up to time $k$, $\{x_0,x_1,...,x_k\}$. Bolded letters $\mathbf{y} = \mathbf{Kx}$ denote infinite horizon signals, and linear convolution operators, such that $y_k = \sum_{t=0}^k K_t x_{k-t}$. The norm operator $\norm{\cdot}$ is overloaded to apply to signals and convolution operators: $\norm{\mathbf{x}} = \sup_k \norm{x_k}$ and $\norm{\mathbf{\Phi}} = \sup_{||\mathbf{w}||\leq 1} \norm{\mathbf{\Phi w}}$. The norm on linear operators $\norm{\mathbf{\Phi}}$ is also the induced norm. If unmarked, the norm is assumed to be the $\infty$-norm.
\subsection{LTI System with Perception Model}
We consider a LTI system, following the setup in \cite{Dean2019a}:
\begin{align}\label{eq:system}
x_{k+1} &= Ax_k + Bu_k +Hw_k \\
z_k &= q(x_k) \\
y_k &= p(z_k) = Cx_k + e_k \label{eq:err_model} \\
u_k &= \mathbf{K}(y_{0:k})\label{eq:K}
\end{align}

In this system, the state at time $k$ is $x_k \in \R^6$, the input is $u_k \in \R^3$, the high-dimensional pixels and IMU measurement is $z_k$, the measurement after utilizing the perception map is $y_k \in \R^6$, and the system is perturbed by noise $w_k$ with $||H||=1$.  The matrices $A,B, C,$ and $H$ are all known and described in Section \ref{sec:quadmodel}. The linear time invariant output feedback controller $\mathbf{K}$ is described later in this section.


The high-dimensional measurement $z_k$ represents the raw sensors' measurements which depend on the system's state according to an observation process $q(x_k)$. In a VIO system, $z_k$ corresponds to raw images and IMU measurements. The VIO algorithm takes $z_k$ as an input and outputs lower dimensional state measurement $y_k$, according to the VIO perception map $p(z_k)$.  We model the output of the perception map $p(z_k)$ as being composed of two terms: the first, $Cx_k$, corresponds to an idealized sensor output (e.g., the full system state), whereas the second, $e_k := p(z_k) - Cx_k$, is a state-dependent error term.  With this perception model in place, we can ignore the intermediate measurement $z_k$, resulting in a direct noisy observation model $y_k = Cx_k + e_k$, as VIO systems directly provide an approximate system state measurement. 
At a high level, our goal is to ensure that the system response to perception errors $e_k$ remains bounded. System Level Synthesis (SLS) \cite{sls_tutorial} provides a framework for doing so. SLS provides a parameterization of the achievable closed-loop behaviors of the LTI system \eqref{eq:system} under the feedback policy $\mathbf{u}=\mathbf{K}\mathbf{y}$ that makes explicit the effects of process noise $H\mathbf{w}$ and measurement errors $\mathbf e$ on system behavior.
This means that we can write the state and input as result of a convolution of the state noise, perception noise and the closed-loop system responses $\{\Phi_{xw}(k), \Phi_{xe}(k), \Phi_{uw}(k), \Phi_{ue}(k)\}$,
\begin{align}\label{eq:phis_discrete}
\begin{bmatrix} x_k \\ u_k \end{bmatrix} = \sum_{t=1}^{k} \begin{bmatrix} \Phi_{xw}(t) & \Phi_{xe}(t) \\ \Phi_{uw}(t) & \Phi_{ue}(t)\end{bmatrix} \begin{bmatrix} Hw_{k-t} \\ e_{k-t}\end{bmatrix} \:,
\end{align}
which we can compactly represent in the $z$-domain as:
\begin{align}\label{eq:phis}
\begin{bmatrix}
\mathbf{x} \\ \mathbf{u}
\end{bmatrix}
 =
  \begin{bmatrix}
  \mathbf{\Phi_{xw}} & \mathbf{\Phi_{xe}} \\
  \mathbf{\Phi_{uw}} & \mathbf{\Phi_{ue}}
  \end{bmatrix}
  \begin{bmatrix}
  H \mathbf{w} \\
  \mathbf{e}
  \end{bmatrix}
\end{align}
In \cite{sls_tutorial}, it is shown that to characterize all possible system responses achievable by a stabilizing feedback controller $\mathbf K$, it is necessary and sufficient to constrain these system responses to be stable, and satisfy the affine constraints
\begin{equation}\label{eq:sls_constraints}
\begin{matrix}
    \begin{bmatrix}
    zI-A & - B
    \end{bmatrix}
    \begin{bmatrix}
    \mathbf{\Phi_{xw}} & \mathbf{\Phi_{xe}} \\
    \mathbf{\Phi_{uw}} & \mathbf{\Phi_{ue}}
    \end{bmatrix}
    = \begin{bmatrix}
    I & 0
    \end{bmatrix}
  \\[.4cm]
    \begin{bmatrix}
    \mathbf{\Phi_{xw}} & \mathbf{\Phi_{xe}} \\
    \mathbf{\Phi_{uw}} & \mathbf{\Phi_{ue}}
    \end{bmatrix}\begin{bmatrix}
    zI-A \\ - C
    \end{bmatrix}
    = \begin{bmatrix}
    I \\ 0
    \end{bmatrix}.
    \end{matrix}
\end{equation}
Then, given any system responses satisfying these constraints, the controller $\mathbf{K} = \mathbf{\Phi_{ue}} - \mathbf{\Phi_{uw}}\mathbf{\Phi_{xw}}^{-1}\mathbf{\Phi_{xe}}$ will achieve the desired behavior in system \eqref{eq:system}. We point the reader to \cite{sls_tutorial} for a thorough explanation of SLS, and Section 4.1 of \cite{Dean2019a} for a more targeted overview of SLS as applied to the perception based problem considered in this paper.

From \eqref{eq:phis}, we see that $\mathbf{x} = \mathbf{\Phi_{xw}} H \mathbf{w} +  \mathbf{\Phi_{xe}} \mathbf{e} $. Thus, the closed loop map from perception error $\mathbf{e}$ to state $\mathbf{x}$ is $\mathbf{\Phi_{xe}}$. Intuitively, the size of the system response $\norm{\mathbf{\Phi_{xe}}}$ captures how aggressively the system can react to the perception error $\mathbf{e}$ -- therefore, in order to bound the system's response to perception errors, we need to bound $\norm{\mathbf{\Phi_{xe}}}$. This bound will be influenced by the characteristics of $\mathbf{e}$; if $\mathbf{e}$ is large, $\norm{\mathbf{\Phi_{xe}}}$ must be small, roughly speaking. As such, we first characterize an error model $\mathbf{e}$ according to \eqref{eq:err_model}. Specifically, using ground truth data measurements $y_k$ and deriving $C$ from the system, we can then characterize the nonlinear error $\norm{\mathbf{e}} = \norm{p(\mathbf{z}) - C\mathbf{x}}$ from training data. Then, we use the properties of $\mathbf{e}$ to constrain $\norm{\mathbf{\Phi_{xe}}}$ in the controller optimization problem. This constraint is the \textit{robustness constraint} that endows the resulting controllers with robustness guarantees, as described at the end of this section.

\subsection{Data-Driven Error Model \& Robust Synthesis}

In order to craft a robustness constraint, we need to characterize our perception error model $\mathbf{e}$. Our assumption of a static observation model $z_k = q(x_k)$ implies that the perception error model fit on training data remains valid if those states are visited again at test time.   Under suitable smoothness assumptions on the observation model $q$ and perception map $p$, we expect the error model to be approximately valid on states near to those visited during training. 
But how close do we need to stay in order for our perception error model to be approximately valid, and what rate does its accuracy degrade?
The answer from \cite{Dean2019a} is to quantify the so-called $S$-slope of the error model in a neighborhood of the training data to quantify this degradation. One can think of the $S$-slope as a worst-case bound on the rate at which the perception error model degrades as the system moves away from states found in the training data. Formally, we define the data-driven estimate $\hat{S}$ of the $S$-slope with radius $r$ around the training data $\{x_1,...,x_n\}$ as:
\begin{align}
\hat{S} &= \max_{x_i \in \mathcal{S}_{x_d}} \frac{\norm{e(x_i) - e(x_d)}}{\norm{x_i - x_d}}\\
 \mathcal{S}_{x_d} &= \{x_1,..., x_n : \norm{x_i - x_d} < r \} 
\end{align}

It is shown in \cite{Dean2019a} that using both the $S$-slope and a norm bound $\norm{\mathbf{e}}$ of the error model, a robustness constraint can be added to the control synthesis problem to guarantee a bounded perception error. In particular, we utilize the SLS problem formulation for a \rpb from \cite{Dean2019a}:
\begin{align}
  \min_{\mathbf{\Phi}} \quad &
 c(\mathbf{\Phi_{xw}}, \mathbf{\Phi_{uw}}, \mathbf{\Phi_{xe}}, \mathbf{\Phi_{ue}}) \label{eq:sls} \\
  \textrm{s.t.} \quad & \text{Equation \eqref{eq:sls_constraints},}\\
  &\norm{\mathbf{\Phi_{xe}}} < \frac{1 - \frac{1}{r} \norm{\mathbf{\Phi_{xw}} H \mathbf{w} - \mathbf{x_d}}}{S + \frac{\eps_e}{r}}    \label{eq:robustness_constraint}
\end{align}
from which we obtain the system level responses $\{\mathbf{\Phi_{xw}}, \mathbf{\Phi_{xe}}, \mathbf{\Phi_{uw}}, \mathbf{\Phi_{ue}}\}$, and can construct the controller $\mathbf{u} = \mathbf{K} \mathbf{y}$, with $\mathbf{K} = \mathbf{\Phi_{ue}} - \mathbf{\Phi_{uw}} \mathbf{\Phi_{xw}}^{-1} \mathbf{\Phi_{xe}}$. We note that the formulation shown here is for an infinite horizon time response. In practice we utilize a Finite Impulse Response approximation of this procedure, outlined in \cite{sls_tutorial} and \cite{Dean2019a}.

As described in Theorem 4.2 from \cite{Dean2019a}, the resulting controller guarantees that the system remains within a bounded distance of states visited during training, and consequently, we can guarantee that our perception errors $\norm{\mathbf{e}} = \norm{p(\mathbf{z}) - C \mathbf{x}}$ also remain bounded:
\begin{align}\label{eqn:generalization_guarantee}
\norm{\mathbf{e}} = \norm{p(\mathbf{z}) - C \mathbf{x}} \leq
\gamma := \frac{\norm{\hat{\mathbf{x}} - \mathbf{x}_d}+R_0}{1-S\norm{\mathbf{\Phi_{xe}}}}
\end{align}

Our goal in this work is to demonstrate the practical relevance of this robustness guarantee in a quadrotor system.
\subsection{Imitation Cost Function}
In order to synthesize a controller, we need a cost function. A typical choice is the quadratic cost $c_{QR}$ with state cost weights $Q$ and input cost weights $R$ \cite{Dean2019a}. The subsequent controller is called LQG or $L_1$ controllers, depending on choice of norm in the cost function ($L_2$ or $L_\infty$):
\begin{align}\label{eq:qrcost}
\begin{split}
c_{QR}(&\mathbf{\Phi_{xw}}, \mathbf{\Phi_{uw}}, \mathbf{\Phi_{xe}}, \mathbf{\Phi_{ue}})\\ =
  &\norm{\begin{bmatrix}
    Q^{1/2} & \\
    & R^{1/2}
    \end{bmatrix}
  \begin{bmatrix}
  \mathbf{\Phi_{xw}} & \mathbf{\Phi_{xe}} \\
  \mathbf{\Phi_{uw}} & \mathbf{\Phi_{ue}}
  \end{bmatrix}
  \begin{bmatrix}
  \varepsilon_w H \\ \varepsilon_e I
  \end{bmatrix}
  }
  \end{split}
\end{align}
One problem with this cost function is that if the controller need to be adjusted for better performance, it can be difficult to tune this cost function with $Q$ and $R$. While there are formal rules of thumb in crafting $Q$ and $R$, small changes can result in wildly different performance. Additionally, the difference between the quadratic cost controller with and without the robustness constraint \eqref{eq:robustness_constraint} can be significant.

In this work, we introduce a new cost function called \textit{imitation cost} $c_{im}$. The intuition behind this cost is that for most applications of robust control, an existing, non-robust controller already exists and works well. We do not want to synthesize a controller with completely different behavior from this nominal controller, but instead would simply like to have a controller which "imitates" the nominal controller, while still being robust. Hence, the cost should be to minimize the difference between the robust controller and the desired controller while subject to the robustness constraint \eqref{eq:robustness_constraint}.

We define the nominal controller system responses as $\mathbf{\Phi_{xw, n}, \Phi_{uw, n}, \Phi_{xe, n}, \Phi_{ue, n}}$, and still retain the quadratic cost structure such that it is a convex program. In practice, we find this controller much easier to "tune", since we may simply tune our nominal controller, which could be as simple and intuitive as a PD controller, and then impose the robustness constraint afterwards through optimization. For this cost, we found the $L_1$ yielded the best performance.
\begin{align}\label{eq:immitationcost}
\begin{split}
c_{im}(&\mathbf{\Phi_{xw}}, \mathbf{\Phi_{uw}}, \mathbf{\Phi_{xe}}, \mathbf{\Phi_{ue}})\\
  =&\norm{\begin{bmatrix}
    Q^{1/2} & \\
    & R^{1/2}
    \end{bmatrix}
  \begin{bmatrix}
  \mathbf{\Phi_{xw} - \Phi_{xw, n}} & \mathbf{\Phi_{xe}- \Phi_{xe, n}} \\
  \mathbf{\Phi_{uw}- \Phi_{uw, n}} & \mathbf{\Phi_{ue}- \Phi_{ue, n}}
  \end{bmatrix}
  }
 \end{split}
\end{align}
\subsection{Linear Quadrotor Model}\label{sec:quadmodel}
In order to use the results from \cite{Dean2019a}, we require an LTI model of our quadrotor. We assume an underlying attitude controller as described in \cite{Mellinger2011}, and that we are not flying aggressively. Thus we can linearize our quadrotor around hover while holding yaw constant to get an LTI model.

We define the states of the quadrotor to be the position and velocity of the center of mass: $\mathbf{x} = [x\ y\ z\ \dot{x}\ \dot{y}\ \dot{z}]^T$. The inputs for the quadrotor are total propeller thrust and instantaneous attitude in Euler angles as $\psi, \theta$ and $\phi$, or yaw, pitch and roll. Yaw is held constant at $\psi = 0$, and thus dropped from the input space. Thus, our final control inputs are pitch, roll, and total thrust or $\mathbf{u} = [\theta\ \phi\ f_t]^T$.

The quadrotor system is linearized around hover as $\bar{\mathbf{x}} = [0\ 0\ 0\ 0\ 0\ 0 ]^T$, $\bar{\mathbf{u}} = [0\ 0\ mg]^T$, where $g$ is acceleration due to gravity and $m$ is the mass of the quadrotor. Finally, we can describe our continuous LTI system:
\begin{align}
\mathbf{\dot{x}} &= A\mathbf{x} + B\mathbf{u} \\
A &=
\begin{bmatrix}
0 & 0 & 0 & 1 & 0 & 0 \\
0 & 0 & 0 & 0 & 1 & 0 \\
0 & 0 & 0 & 0 & 0 & 1 \\
0 & 0 & 0 & 0 & 0 & 0 \\
0 & 0 & 0 & 0 & 0 & 0 \\
0 & 0 & 0 & 0 & 0 & 0
\end{bmatrix}
B = \begin{bmatrix}
0 & 0 & 0 \\
0 & 0 & 0 \\
0 & 0 & 0 \\
-g & 0 & 0 \\
0 & g & 0 \\
0 & 0 & \frac{1}{m}
\end{bmatrix}
\end{align}
The perception map of our system $p(z_k)$ from VIO returns measurements of position and velocity, so $C$ is simply the identity:
\begin{align}
y_k &= p(z_k) = Cx_k + e_k  \\
C &= I_{6 \times 6}
\end{align}
To match both our simulation and hardware setups, we convert our system to discrete time with $\Delta T = 0.1 $ seconds.
\section{Experimental Setup}
\comment{In this section, we investigate our \rpb in comparison with other controllers in trajectory following, on both simulation and a hardware platform. The testbed robot is ... which features VIO as the state estimator. For our controller, the data generation to train the error model is collected by flying the robot using Mellinger et al. approach~\cite{Mellinger2011}.}
Our experiments consist of flying around a trajectory near hover, with access to both ground truth and real state estimation of our quadrotor system. To generate the desired trajectory for the position controller, we utilize a time-parameterized circular trajectory at a constant height. For state estimation, we use VIO which functions as the perception map. In the experiment, we first fly our trajectory with a non-linear $SO_3$ controller from \cite{Mellinger2011} in order to gather data to train the perception error model. From the error model, we synthesize controllers using SLS, then fly the same circle trajectory to compare results.
\subsection{Simulation Platform}
ROS and Gazebo function as the simulation environment for a World Electronics Dragon Drone Development quadrotor. To simulate the visual perception, we utilize a Multi-state Constraint Kalman Filter with Visual-Inertial Odometry (MSCKF-VIO) \cite{Sun2018}, which is integrated with simulated IMU and stereo camera signals from Gazebo in order to produce a VIO measurement. We designed this simulation platform to best mimic real platforms while still having access to perfect ground truth in order to verify our guarantees.


\subsection{Hardware Platform}
\begin{figure}[thpb]
	\centering
	\parbox{3in}{\centering
	\includegraphics[width=0.7\columnwidth]{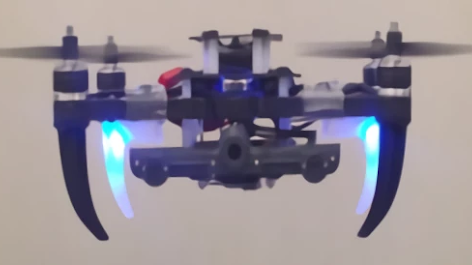}}
	\caption{ The Snapdragon platform we used in our hardware experiments. It features a downward facing camera and an on-board VIO state estimator. }
	\label{fig:snapdragon}
\end{figure}

The quadrotor platform is based on the quadrotor used in \cite{snapdragon}. It is a 0.16 m diameter, 256 g quadrotor using a Qualcomm®, Snapdragon™ Flight™ with a 7.4 V LiPo battery. The hardware platform is depicted in Figure \ref{fig:snapdragon}. The visual measurement system consists of a downward facing camera with a fish-eye lens and 107$^\circ$. field of view.
The downwards facing camera is the main differentiator from the simulated VIO system, but otherwise is very similar to the simulated quadrotor. For the ground truth, we utilize the VICON Motion Capture system.
\section{RESULTS}
In order to generate the robust controllers, we first characterize the error profile of the perception system to generate a \rpb in Section \ref{sec:err_profile}. We then test our controllers in simulation and reality. Unfortunately, only the PD and $L_1$ Robust controllers were sufficiently safe to realize on the hardware platform. We compare controller performance in Sections \ref{sec:sim_tracking_performance} and \ref{sec:hardware_tracking_performance}. State estimation of controllers as well as comparison to the theoretical bound is discussed in Section \ref{sec:bounded_err}.

\subsection{Perception Error Profile}\label{sec:err_profile}
For the error model, we wish to characterize the perception map $p(z_k)$ by fitting an error profile to match $e_k = p(z_k) - C x_k$ using training data. The training data were generated by flying a circular trajectory using a high performance nonlinear $SO_3$ controller. The norm and local $S$-slope bounds $\norm{e_k} \leq \epsilon_e$, and $S$ are utilized in the SLS procedure described in equations \eqref{eq:sls}. The $S$-slope represents how quickly the perception error can change, and thus if $S$ is large, then the controller must be extremely conservative as the perception bias may grow extremely quickly away from training data points.

While theoretically sound, unfortunately, the empirical local S-slope of the error profile was found to be an over-conservative bound of reality. Our characterization of the error profile resulted in $S = 11.7$ in simulation and $S = 14.4$ in real experiments. This resulted in too stringent a robustness constraint, rendering the SLS optimization problem \eqref{eq:sls} infeasible. In reality, only a few points in each data set exhibit such a high $S$-slope - meaning that a few measurements have high noise. This is expected, as there are often "bad measurements" (especially in velocity) which are largely outliers and not representative of the perception bias growing extremely quickly. Therefore, we take our estimate of the $S$-slope for the purposes of our robust controller to be far less conservative such that the SLS procedure is feasible. We further justify it by evaluating in our results how well our robustness guarantee was maintained, even with a less conservative $S$-slope on the error profile, as seen in the next Section \ref{sec:bounded_err}. In future work, we hope to address this problem with a more appropriate perception model.

\subsection{Controllers}
We compared the following four controllers, all of which were synthesized using optimization problem \eqref{eq:sls}, except for the PD controller:
\begin{itemize}
    \item Well-tuned PD controller. This controller was tuned to track very well under ideal perception conditions.
    \item Nominal $L_1$ controller. This controller was synthesized using the cost function from equation \eqref{eq:qrcost} with the $L_1$ norm, and without enforcing the robustness constraint \eqref{eq:robustness_constraint}. We choose to use this instead of a Linear Quadratic Gaussian due to superior performance. The only difference is the use of the $\infty\to\infty$-induced-norm or Frobenius norm in the cost of the controller synthesis problem.
    \item $L_1$ Robust controller with the normal cost. This controller was synthesized using the cost function from equation \eqref{eq:qrcost}.
    \item $L_1$ Robust imitation controller. This controller was synthesized using the cost function from equation \eqref{eq:immitationcost}. The nominal controller it was imitating is the $L_1$ controller.
\end{itemize}
In our hardware experiments, only the $x-y$ position control of the $L_1$ robust imitation controller was stable enough to test against the PD controller. The $z$ control of the PD controller was used instead of the $z$ control for the $L_1$ robust imitation controller for safety. We have matched our simulated data to also use the $z$-control of the PD controller for a more direct comparison.


\subsection{Simulated Controller Performance}\label{sec:sim_tracking_performance}
\begin{figure}[thpb]
      \centering
      \parbox{3in}{
      \includegraphics[width=0.8\columnwidth]{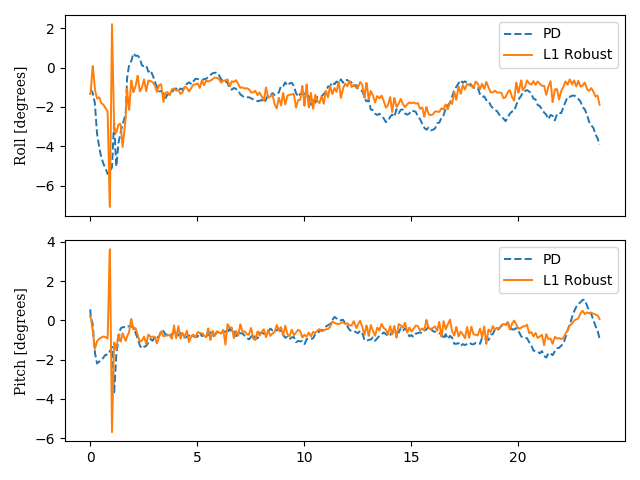}}
      \caption{The control input on a L1 robust imitation control run compared to what the PD control from the same state estimate.}
    \label{fig:u_L1_pd}
\end{figure}

The trajectory tracking was done around a circular trajectory. The simulated resulting position tracking error can be seen in Figure \ref{fig:pos_tracking}, while the estimation errors can be seen in Figure \ref{fig:est_err}. In the simulated experiments, the controllers all tracked the trajectory, with the PD controller tracking the best. Qualitatively, the robust controllers reacted more conservatively to errors in state estimation, as seen in Figure \ref{fig:u_L1_pd}. We see that at the beginning of the trajectory between 0-4 seconds, there is a disturbance in the state estimation to which the $L_1$ robust controller reacts quickly, as we can see with the spike, but quickly damps out to a more conservative control than the PD controller. This results in a "softer" controller overall, but is important in real-life when erroneous measurements make their way into the control. Indeed, there is a trade-off in how aggressively a controller can react to changes in sensor measurements, and how robust it can be to perception errors.  Our robust controller gives a principled way to navigate this trade-off, allowing us to balance between over-reacting to perception errors and maintaining good tracking.

\subsection{Hardware Controller Performance}\label{sec:hardware_tracking_performance}
\begin{figure}[thpb]
      \centering
      \parbox{3in}{
      \includegraphics[width=0.9\columnwidth]{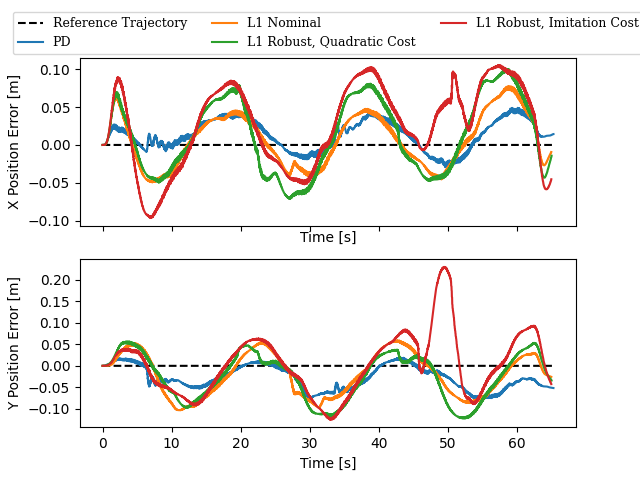}
     \includegraphics[width=0.9\columnwidth]{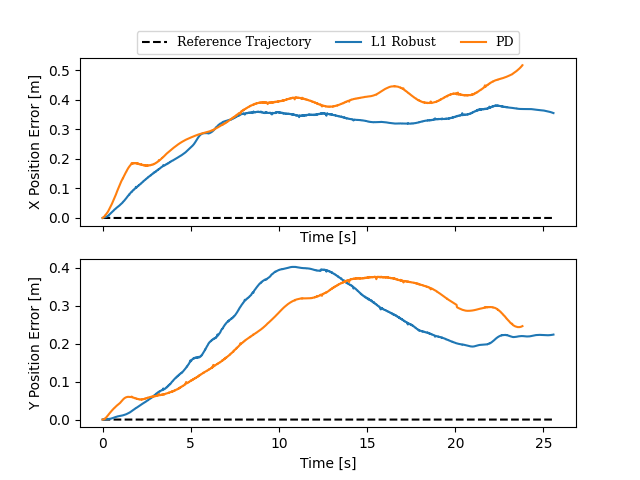}}

      \caption{The position tracking of the controllers around the circle trajectory. (Top) The controllers in simulation did three laps, and all controllers perform similarly well.\\
      (Bottom) Position tracking of the $L_1$ Robust imitation controller and PD controller on the hardware system. Both had fairly similar performance qualitatively, though the PD controller tracked slightly closer than the $L_1$ robust controller.}
    \label{fig:pos_tracking}
\end{figure}

The position tracking in the hardware experiments is shown in Figure \ref{fig:pos_tracking}. The PD controller and $L_1$ controller are on par in tracking performance, however, the PD control looks slightly smoother. The perception error is also about the same for both controllers, as seen in Figure \ref{fig:est_err}.
\subsection{Sources of Difficulty in Controller Performance}
We suspect the difficulty in synthesizing the optimal controllers for the real quadrotor system lies in the fact that the quadrotor is linearized in hover, but is not truly a linear system. In the simulation, where process noise is closer to Gaussian, the controllers are able to stabilize the system as expected. However, in real life, the process noise may be more due to model error, which may interfere with the synthesized controllers in a pathological way.

For the simulation, we still experienced significant difficulty in tuning controllers to be performant enough to satisfy the task. The synthesized controllers often let the quadrotor drift farther from the desired trajectory than the PD controller. Only after much tuning did we achieve tracking error similar to PD as seen in Figure \ref{fig:pos_tracking}. We hypothesize that with a more stable system, the robust controllers as well as nominal $L_1$ controllers would be adequate controllers. However, even with our generous interpretation of the error profile $S$-slope, the trade-off between minimizing perception error and minimizing tracking error is too far in the direction of minimizing perception error. Ultimately, the quadrotor needs to track well enough to perform its task. Qualitatively speaking, this highlights a deficiency in this \rpb formulation that the robustness constraint gives a hard limit on how much the roboticist tune her preference for minimizing perception error or tracking error while remaining robust. The "tuning knobs" available in the cost function given in equation \eqref{eq:qrcost}  are the $Q$ and $R$ matrices, which in our experience was insufficient to produce a controller good enough to use on a real hardware platform.


\subsection{Bounded Perception Error Guarantee}\label{sec:bounded_err}
The first guarantee in equation \eqref{eqn:generalization_guarantee} simply states that our state estimation error will remain bounded. We can confirm this directly in simulation and reality by comparing our ground truth (simulation or VICON motion capture) to the VIO state estimate.

\begin{figure}[thpb]
	\centering
	\includegraphics[width=0.85\columnwidth]{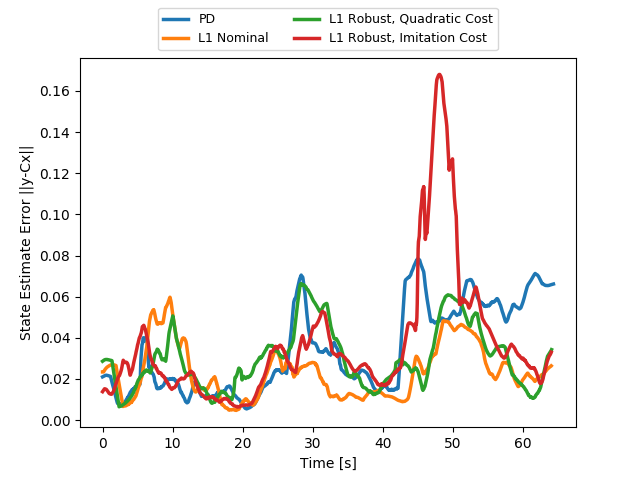}
	\includegraphics[width=0.85\columnwidth]{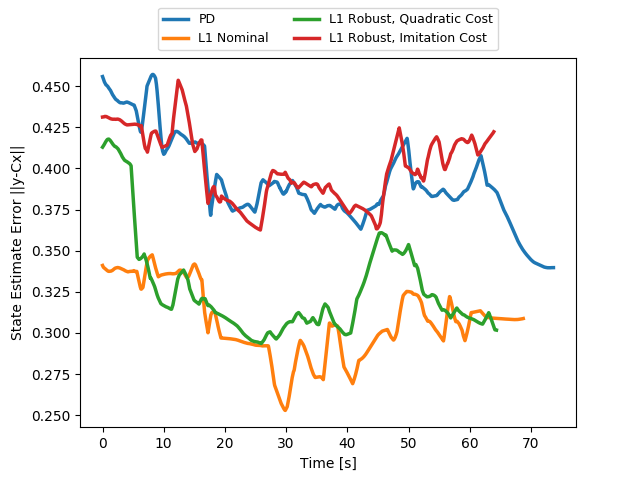}
	\includegraphics[width=0.78\columnwidth]{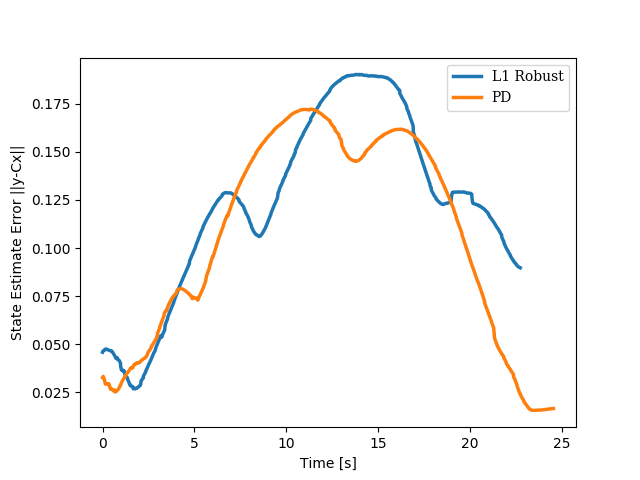}

	\caption{The norm of the perception error over time from simulation with regular perception (Top), simulation with degraded perception (Middle) and our hardware experiments (Bottom). The perception error stays below the theoretical bound for all controllers. The signals has been averaged with a time constant of 1 second for readability.}
	\label{fig:est_err}
\end{figure}

We compare the perception error to the theoretical bound in equation \eqref{eqn:generalization_guarantee}, with a slight modification. We compare instead to a slightly stricter bound, which removes the dependence on the distance to the training data $\norm{\hat{\mathbf{x}} - \mathbf{x}_d}$, which is difficult to calculate.
\begin{align}\label{eqn:strong_generalization_guarantee}
\norm{p(\mathbf{z}) - C \mathbf{x}} \leq \frac{R_0}{1-S\norm{\mathbf{\Phi_{xe}}}}
\end{align}
In this case, the perception error still stays below the theoretical limit, for all controllers, in simulation and reality, as seen in Figures \ref{fig:est_err}. The numerical values for these limits are $1.602$ in simulation and $3.172$ in our hardware experiments, calculated using the true $S$-slope of the true error profiles. We note that these are an order of magnitude larger than the errors we see in either simulation or reality. While our controllers satisfy these theoretical bounds, they are hardly reassuring to consider that a quadrotor may drift by almost a meter in position (since most of the state estimation error bias comes from position), which is not acceptable for many applications.

Additionally, we performed the same long-trajectory tests with increased noise in the virtual camera sensor, which further highlighted the difference between the naive and robust controller performance. The perception error in these trials can be seen in Figure \ref{fig:est_err}. In this case, the $L_1$ quadratic cost robust and nominal controllers performed significantly better than the PD and $L_1$ robust imitation cost controllers. This suggests that the quadratic cost controllers may perform significantly better than controllers tuned for ideal perception conditions. In particular, the $L_1$ robust controller with quadratic cost appears to reduce the perception error the most. However, once again, all of the controllers stay far below the perception error theoretical bound.
\section{Discussion}
In this work, we implemented a \rpb on a quadrotor platform. With this controller design, the control is aware of the perception error, and drives the quadrotor state to keep the perception error bounded. We implement the first version of a \textit{robust}, perception-based controller on a simulated and hardware quadrotor platform. In implementing and testing the control synthesis procedure with a quadrotor perception system, we encountered issues in the controller design relying on our perception map being unrealistically smooth, causing the control synthesis problem to be infeasible. In reality, using a slightly less conservative characterization of the perception map resulted in a feasible but still empirically robust perception-based controller.

We also introduced a new cost function to aid with tuning the \rpb. It takes an existing good controller and "robustifies" it with the SLS procedure, which in both simulation and hardware we found to be a more reliable way to produce good controllers.

We tested the \rpb against a well-tuned PD controller and $L_1$ controller and saw that the \rpb and PD controller performed equally well, significantly better than the $L_1$ controller. Both PD and the \rpb satisfied the robustness constraint.

In degraded perception conditions mimicking adverse lighting conditions, the \rpb adjusted its control to maintain better perception. Overall, we found that the robust control was overly conservative for well-lit conditions, and only for degraded perception does its advantages come into play. This suggests further experimentation in more challenging state estimation conditions.
\section{Future Work}
Moving forward, we would like to explore new controller formulations using System Level Synthesis which do not rely upon overly-conservative characterizations of perception noise such as $S$-slope. Instead, we believe interesting work lies in characterizing parts of the state space as having better or worse state estimation, and using that information to formulate a robust, perception aware controller. We would also like to incorporate a linear time-varying (LTV) version of the robust perception-based control synthesis, based on \cite{Wang2019}. This will allow us to incorporate yaw back into the controller for the quadrotor, which is important for forward-facing camera scenarios and would be a robust complement to current research that investigates active sensing with yaw. We also see promise in maintaining the robustness constraint. This would be applicable in patrol or inspection situations, where a vehicle needs to be robust but still operate in a noisy environment. The robust controller could also be used in combination with a safe exploration policy, where reliance on perfect perception is unrealistic, yet safety and robustness are still needed.

\section*{ACKNOWLEDGMENTS}
We would like to acknowledge Dinesh Thakur and Alex Zhou for their help in using the Snapdragon Quadrotor platform and the VICON Motion capture system, and Ke Sun for his help with the simulated perception.

This work was supported in part by the Semiconductor Research Corporation (SRC) and DARPA.

\addtolength{\textheight}{-12cm}   

\bibliographystyle{ieeetr}
\bibliography{bib, other_bib}

\end{document}